\pdfoutput=1
\documentclass[11pt]{article}
\usepackage[final]{acl}
\usepackage{amsmath}
\usepackage{times}
\usepackage{latexsym}
\usepackage[breakable]{tcolorbox}
\usepackage[T1]{fontenc}
\usepackage[utf8]{inputenc}
\usepackage{microtype}
\usepackage{inconsolata}
\usepackage{graphicx}
\usepackage{float}
\usepackage{array}
\usepackage{booktabs}
\usepackage{tikz}
\usepackage{colortbl}
\usepackage{makecell}
\usepackage{xcolor}
\usepackage{enumitem}
\definecolor{darkgreen}{rgb}{0, 0.4, 0} 
\definecolor{lightgray}{gray}{0.95} 

\newcommand{\scorechange}[3]{%
  \tikz[baseline=(base.base)]{
    \node[inner sep=0, outer sep=0] (base) {#1};
    \node[anchor=west, draw=none, rectangle, rounded corners=3pt, inner sep=1.5pt, fill=#3!20, font=\sffamily\tiny] (change) at ([xshift=0.3ex, yshift=-0.2ex]base.north east) {#2};
  }%
}

\newcolumntype{M}[1]{>{\centering\arraybackslash}m{#1}}
\title{DecorateLM: Data Engineering through Corpus Rating, Tagging, and Editing with Language Models}

\author{
 \textbf{Ranchi Zhao\textsuperscript{1}\thanks{Equal contribution.}},
 \textbf{Zhen Leng Thai\textsuperscript{2}\footnotemark[1]},
 \textbf{Yifan Zhang\textsuperscript{1}\footnotemark[1]},
 \textbf{Shengding Hu\textsuperscript{2}\footnotemark[1]},
\\
 \textbf{Yunqi Ba\textsuperscript{1}},
 \textbf{Jie Zhou\textsuperscript{1}},
 \textbf{Jie Cai\textsuperscript{1}},
 \textbf{Zhiyuan Liu\textsuperscript{2}\thanks{Corresponding author.}},
 \textbf{Maosong Sun\textsuperscript{2}\footnotemark[2]},
\\
 \textsuperscript{1}Modelbest Inc,
 \textsuperscript{2}Department of Computer Science and Technology, Tsinghua University
\\
\{ranchizhao,thaizhenleng123,yifanzhang634,shengdinghu\}@gmail.com
}
\begin{document}
\maketitle
\begin{abstract}
The performance of Large Language Models (LLMs) is substantially influenced by the pretraining corpus, which consists of vast quantities of unsupervised data processed by the models. Despite its critical role in model performance, ensuring the quality of this data is challenging due to its sheer volume and the absence of sample-level quality annotations and enhancements. In this paper, we introduce \textit{DecorateLM}, a \underline{d}ata \underline{e}ngineering method designed to refine the pretraining \underline{co}rpus through data \underline{ra}ting, \underline{t}agging and \underline{e}diting. Specifically, \textit{DecorateLM}  rates texts against quality criteria, tags texts with hierarchical labels, and edits texts into a more formalized format. Due to the massive size of the pretraining corpus, adopting an LLM for decorating the entire corpus is less efficient. Therefore, to balance performance with efficiency, we curate a meticulously annotated training corpus for \textit{DecorateLM} using a large language model and distill data engineering expertise into a compact 1.2 billion parameter small language model (SLM). We then apply \textit{DecorateLM} to enhance 100 billion tokens of the training corpus, selecting 45 billion tokens that exemplify high quality and diversity for the further training of another 1.2 billion parameter LLM. Our results demonstrate that employing such high-quality data can significantly boost model performance, showcasing a powerful approach to enhance the quality of the pretraining corpus.
\end{abstract}

\begin{figure*}
    \centering
\includegraphics[width=\textwidth]{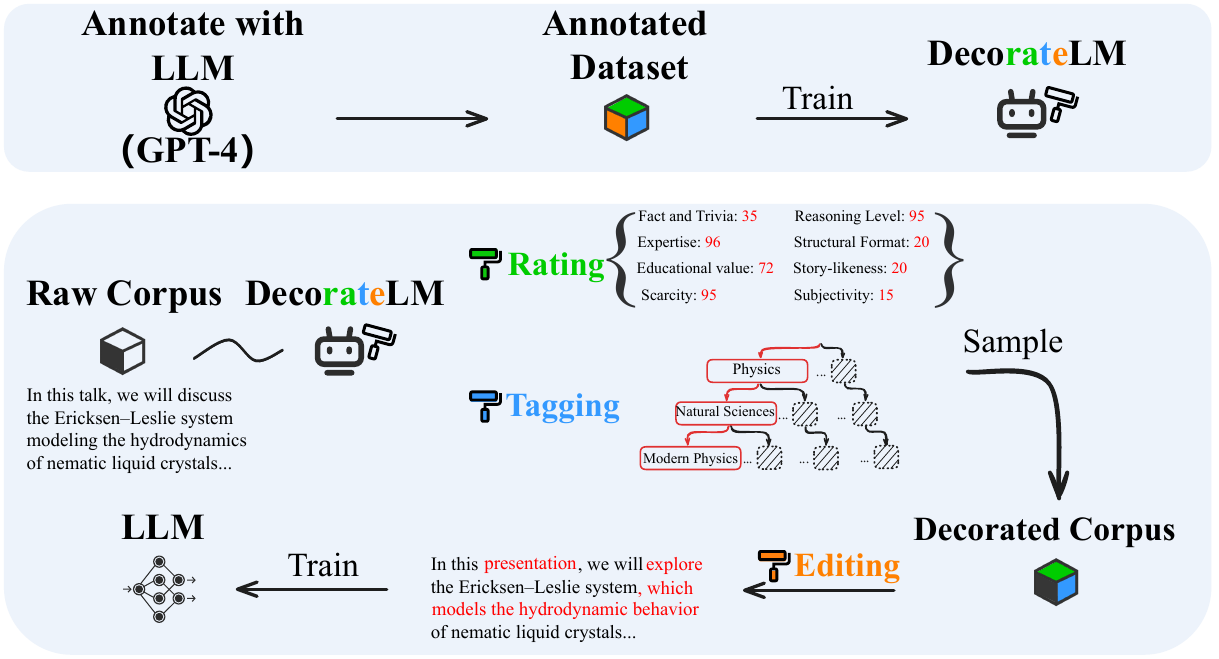}
    \caption{We utilize GPT-4 to assemble an annotated training corpus and integrate data engineering expertise into DecorateLM. DecorateLM is then used to process 100 billion tokens from the raw corpus, sampling 45 billion tokens using its rating and tagging capabilities to create what we refer to as the Decorated corpus. We further enhance the Decorated corpus by applying DecorateLM's editing features, making it more suitable for LLM training.}
    \label{fig:flow}
\end{figure*}

\section{Introduction}
\label{sec:introduction}

\begin{figure*}[t]
    \centering
    \includegraphics[width=\textwidth]{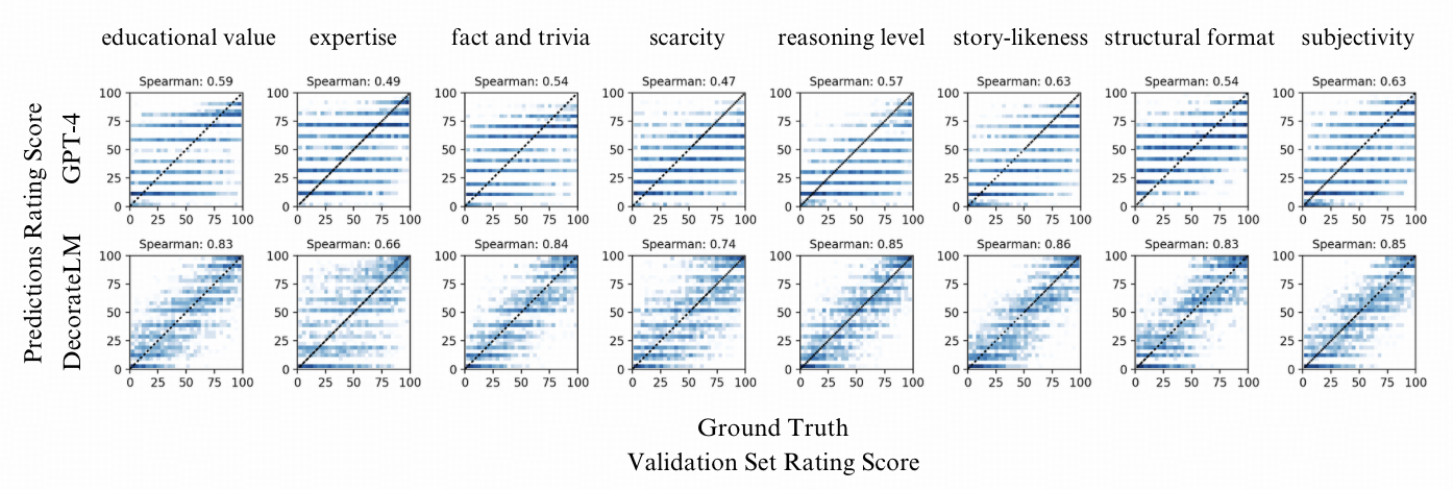}
    \caption{The Spearman correlations between model ratings and ground truth of validation set. Specifically, the x-axis represents the ground truth rating scores of the data. The y-axis represents the prediction rating scores of GPT-4 and DecorateLM after evaluating the validation set. Rating scores generated by GPT-4 are more discrete and inaccurate compared to DecorateLM.}
    \label{fig:spearmanrating}
\end{figure*}

The advent of Large Language Models (LLMs) has ushered in transformative changes across various domains of artificial intelligence~\citep{brown2020language,chowdhery2023palm}, from natural language processing to complex task execution~\citep{qian2023communicative}. The backbone of these models' effectiveness lies in their training processes, specifically in the quality and composition of their pre-training corpora~\citep{penedo2023refinedweb,le2023bloom}. Traditionally, LLMs are pre-trained on vast datasets composed of billions of tokens harvested from diverse text sources.

Data quality is of vital importance for training LLM~\citep{zhou2024lima}. However, acquiring high-quality data is a formidable challenge due to the sheer volume and unstructured nature of it.

The reliance on large-scale unsupervised data leads to the inclusion of numerous low-quality texts within the training data. This infusion of poor-quality data can adversely affect the models' learning processes, resulting in performance deficiencies and limitations in their applicability. However, the existing methods for curating and enhancing the quality of such datasets are often inadequate. They typically lack the capacity to scale to the size required while maintaining or improving data quality, primarily due to the absence of fine-grained annotations and the impracticality of manual oversight.

Addressing these challenges requires innovative approaches that can scale with the data requirements of LLMs while ensuring enhancements in data quality. This paper introduces \textit{DecorateLM}, a comprehensive \underline{d}ata \underline{e}ngineering methodology designed to refine the pretraining \underline{co}rpus through a systematic "decorating" process. The term "decorating" in this context refers to a series of processes aimed at enriching the data with additional metadata, improving its structure, and ensuring its relevance and quality.

\textit{DecorateLM} employs a three-phase strategy to accomplish these goals. The first phase, \underline{ra}ting, involves evaluating texts against a predefined set of quality criteria. 
These criteria are designed to assess the educational value, expertise, fact and trivia, reasoning level, scarcity, structural format, story-likeness and subjectivity of texts.
The second phase, \underline{t}agging, categorizes the texts using a hierarchical label system that reflects the content of the data. This labeling enhances data management and retrieval efficiency, a key aspect of iterative training processes. The final phase, \underline{e}diting, involves revising and standardizing texts to meet higher linguistic standards of formality and clarity.

To implement this methodology effectively, we curate a specialized training corpus using pre-trained LLMs to preprocess and initially rate potential data samples. This approach leverages the model's capabilities to perform initial assessments at scale. We then distill our data engineering expertise into a small language model (SLM)—which is optimized for more detailed and nuanced data processing tasks. We name this SLM as the \textit{DecorateLM}. Using \textit{DecorateLM}, we enhance 100 billion tokens from our initial datasets, selecting 45 billion tokens that exhibit optimal quality and diversity. These tokens are subsequently used to train LM to demonstrate DecorateLM's effectiveness.

The results from our study underscore the substantial benefits of using high-quality, well-curated data in training LLMs. Not only do these results demonstrate improved model performance, but they also suggest that \textit{DecorateLM} offers a scalable and effective solution to one of the most pressing issues in modern AI—enhancing the quality of training datasets amid expanding data requirements. 

\section{Related Work}
\label{sec:relatedwork}

In recent years, the quality and selection of data for training language models receive considerable attention. Researchers propose various methodologies to assess, select, and improve high-quality data, with the goal of enhancing both the performance and efficiency of models~\cite{elazar2023s, longpre2023pretrainer, xie2023data, li2024datacomp}.

\textbf{Data Annotation and Rating}. QuRating, DEITA, and ALPAGASUS are employed for data annotation, each utilizing distinct methodologies to enhance training via refined rating scores~\cite{wettig2024qurating, liu2023makes, chen2023alpagasus}. Phi-1 and MoDS use GPT-4 and DeBERTa to improve educational data and precise data selection, accelerating learning and fine-tuning~\citep{gunasekar2023textbooks, du2023mods}.

\textbf{Domain Diversity in Data}. INSTAG introduces a detailed tagging system for diverse SFT data, improving MT-Bench scores with less data~\citep{lu2023instag}. Phi-1.5 extends Phi-1 by adding synthetic data across multiple domains in a textbook style~\citep{li2023textbooks}.

\textbf{Data Optimization for Model Training}.  
Studies show that models can perform well with smaller datasets and less computing. WRAP maintains performance with fewer resources on the C4 dataset, and TinyStories uses simple vocabulary for quicker learning~\citep{maini2024rephrasing, eldan2023tinystories}. Additionally, Phi-3 uses a two-stage training with web and synthetic data to improve reasoning and specialized skills~\citep{abdin2024phi}.

\section{Method}
\label{sec:headings}
\begin{figure}[t]
    \centering
    \includegraphics[width=\columnwidth]
    {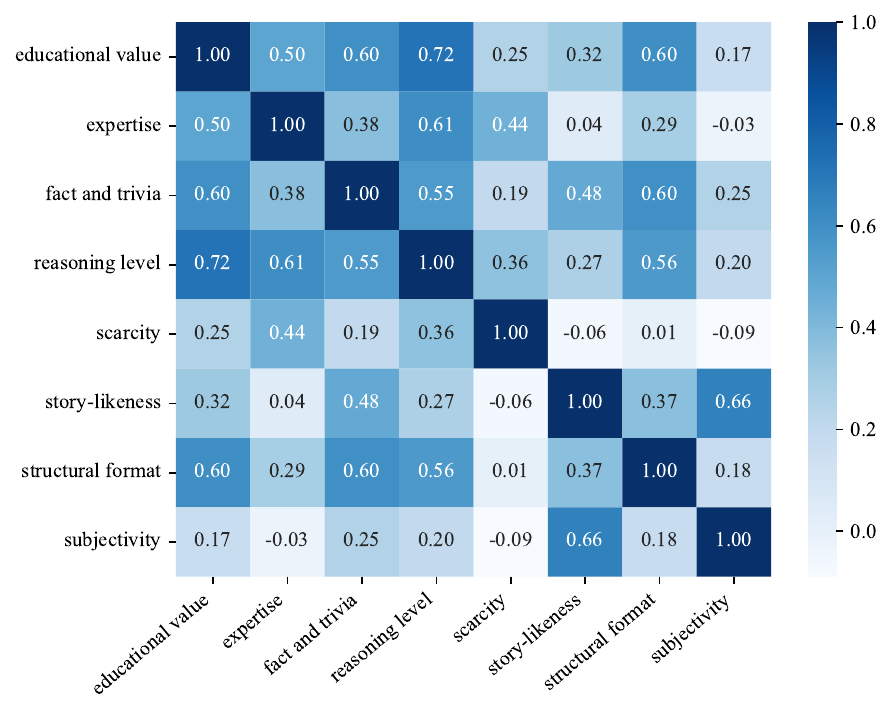}
    \caption{Spearman correlation coefficients between various rating criteria. The correlations align with intuitive expectations. For instance, data with higher educational value often exhibits enhanced reasoning levels, which, in turn, enhances their comprehensibility.
    }
    \label{fig:spearmanscore}
\end{figure}
\subsection{Framework}

In this section, we detail the methodology of DecorateLM, which is designed for sample-level annotation and enhancement. The framework of DecorateLM consists of three distinct phases: rating, tagging, and editing. During the rating phase, DecorateLM assigns numeric scores to a text based on predefined quality dimensions. In the tagging phase, DecorateLM predicts hierarchical tags at three levels for the text. In the editing phase, DecorateLM rephrases the text to present alternative narratives, thereby facilitating the model's acquisition of core knowledge from varied perspectives.

The training pipeline of DecorateLM incorporates both a teacher model and a student model. The teacher model, which is larger, excels in processing detailed instructions related to text quality. However, its slower processing speed limits its practicality for annotating or editing extensive pretraining corpora. To address this, knowledge from the teacher model is distilled into a more compact student model to enhance efficiency. Distinct distillation strategies are employed for each of the three phases. The rating and tagging phases, which involve processing the entire raw corpus and generating concise annotations, exhibit similar input-output dynamics. Consequently, DecorateLM is configured to manage these two phases concurrently to optimize efficiency, instead of leveraging two separate models. For the editing phase, a separate distillation process is implemented to distill the knowledge required for effective rephrasing into another model of DecorateLM.
\subsection{Rating}
\begin{figure}[t]
    \centering
    \includegraphics[width=\columnwidth]
    {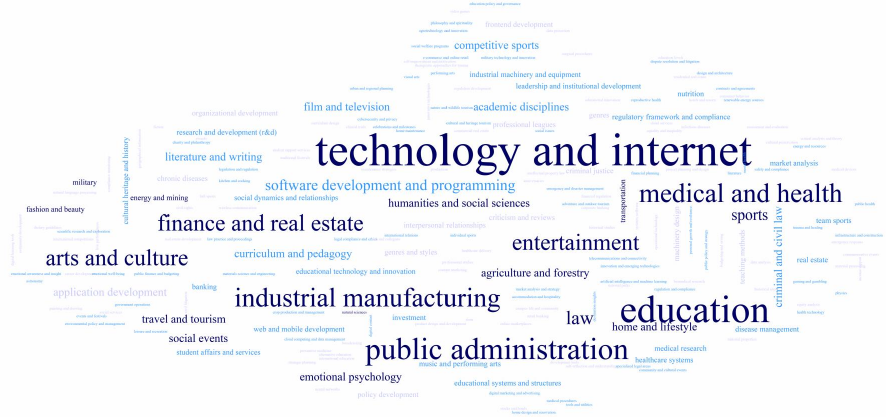}
    \caption{Word cloud of tags. The size of each tag is proportional to its frequency in the annotated dataset. Tags are color-coded based on their levels: first-level tags in dark blue, second-level tags in medium blue, and third-level tags in light blue.}
    \label{fig:wordcloud}
\end{figure}

\begin{figure*}
    \centering
    \includegraphics[width=\textwidth]{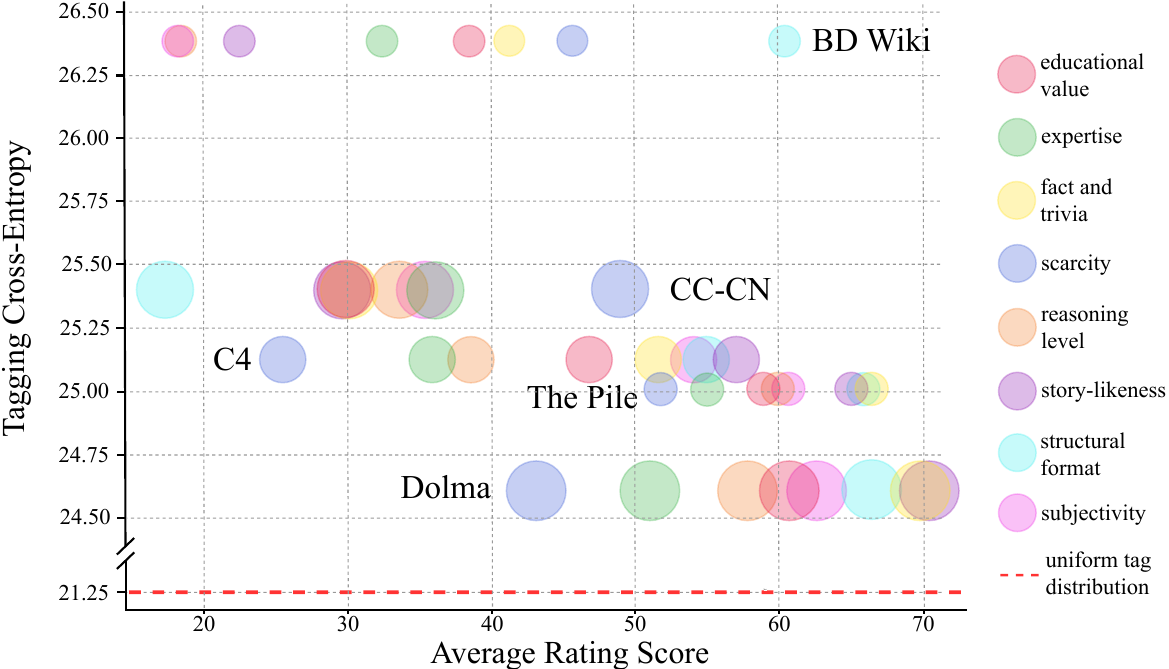}
    \caption{Evaluation of dataset rating and tagging quality using DecorateLM. The x-axis denotes the average rating of each dataset across specified dimensions, whereas the y-axis represents the cross-entropy of tags from predefined tagging system. The circle size correlates with the dataset size.}
    \label{fig:dataset_level_quatlity}
\end{figure*}

High-quality training data is crucial for developing powerful language models. However, the ideal properties that constitute an optimal training corpus remain challenging to characterize comprehensively. To achieve robust language understanding and generation capabilities, language models should be trained on high-quality data meticulously curated based on diverse criteria that capture the essential and abstract qualities of natural language texts. 

\paragraph{Criteria.}
\begin{table}
  \centering
  \begin{tabular}{lccc}
    \toprule
    \textbf{Model}      & \textbf{First} & \textbf{Second} & \textbf{Third} \\
    \midrule
    DecorateLM &         92.1             &       75.6                 &   62.3                    \\
    GPT-4      &          93.6             &        77.3               &          68.5             \\
    \bottomrule
  \end{tabular}
  \caption{\label{tab:taggingacc}
    Comparison of tagging accuracy between DecorateLM and GPT-4 across three hierarchical levels on the validation set. GPT-4, lacking prior knowledge of the designed tagging hierarchy, is provided with the relevant labels for each level through prompts in successive rounds of interaction.
  }
\end{table} 

To assess the quality of texts, we define eight evaluative criteria that quantitatively measure the contributions of a text to model training from multiple perspectives. 
For each criterion, data samples are assigned a quantitative score, enabling an objective evaluation across the various criteria. 

\begin{enumerate}
\item \textit{Educational Value} evaluates whether the content is suitable for educational purposes, specifically its utility in textbooks. It assesses the clarity, detail, and comprehensibility of explanations and guiding principles. 

\item \textit{Expertise} measures the depth of knowledge that content reflects, typically possessed by subject matter experts. 

\item \textit{Fact\&Trivia} focuses on the accuracy of factual information presented in the content, which does not necessarily require specialized expertise to understand. 

\item \textit{Reasoning Level} assesses the necessity for high-level reasoning, sequential thought processes, or chain of thought~\citep{wei2022chain} capabilities in the content. 

\item \textit{Scarcity} targets accurate yet relatively unknown information that is typically familiar only to a select few due to its specialized, niche, or obscure nature.

\item \textit{Structural Format} evaluates the organization and structure of data, such as the use of numbered lists, bulleted lists, and markdown formatting.

\item \textit{Story-likeness} assesses whether the content narrates a story or describes a scenario.

\item \textit{Subjectivity} focuses on content with personal opinions and conversations.


\end{enumerate}

\paragraph{Annotated Dataset Construction.} 
In alignment with the established criteria, we annotate a set of carefully selected samples using GPT-4 to form the annotated dataset. Considering the inaccuracy of LLMs in assigning precise quality scores~\citep{zheng2024judging}, we adopt a pairwise comparison method. Inspired by QuRating~\citep{wettig2024qurating}, this work employs the Bradley-Terry (B-T) model~\citep{bradley1952rank} to derive preference probabilities from pairwise comparisons. All prompts used in the rating phase are displayed in Appendix~\ref{rating prompt}.
Subsequently, we normalize these probabilities by sorting them and applying a linear transformation to map them onto a uniform rating scale from 0 to 100, thereby establishing the final scores for each criterion. 

\paragraph{Analysis.}
Upon acquiring the meticulously curated annotated dataset, we proceed to train DecorateLM, with the training details provided in Appendix~\ref{decoratelm training details rt}. A validation set is segregated prior to training. DecorateLM is employed to assign scores to each data sample. For a fair comparison, we also use GPT-4 to assign numeric scores to these samples. Then we compute the Spearman correlation coefficient between the model-provided scores and the ground truth annotation from the B-T model.
As depicted in Figure~\ref{fig:spearmanrating}, GPT-4, untrained for the rating task, demonstrates \textit{inferior} scoring performance compared to DecorateLM.

In the analysis presented in Figure~\ref{fig:spearmanscore}, we compute the Spearman correlation coefficients between various rating criteria. The results reveal a modest positive correlation across most pairs of criteria, indicating both the independence between different criteria and the commonality present among high-quality texts. 
\subsection{Tagging}
The quality of the pretraining corpus is initially assessed through rating criteria. However, these criteria alone are insufficient for ensuring diversity in the pretraining samples and for the fine-grained selection of data. 
Tagging pretraining data into a broad spectrum of topics and fields can ensure diversity within the training corpus. 
Furthermore, a structured tagging system facilitates the targeted enhancement of the model by incorporating data that address specific areas, consequently improving the model's performance in particular domains. Next, we introduce our hierarchical tagging system.

\paragraph{Tags Design.}
To systematically categorize the pretraining dataset, we first clearly define 21 primary categories that cover a wide range of human knowledge, from Natural Sciences to Social Events. We then expand this framework by engaging GPT-4, which serves as a human expert, in a two-step iterative dialogue process. 
The first dialogue iteration yields 255 second-level tags.
For the third-level tags, we inform GPT-4 of each first-level category along with its corresponding second-level tags, prompting the model to generate a total of 793 specific third-level tags under the second-level categories. 
The details and prompts are in Appendix~\ref{tagging prompt}.


\begin{figure}[t]
    \centering
    \includegraphics[width=\columnwidth]{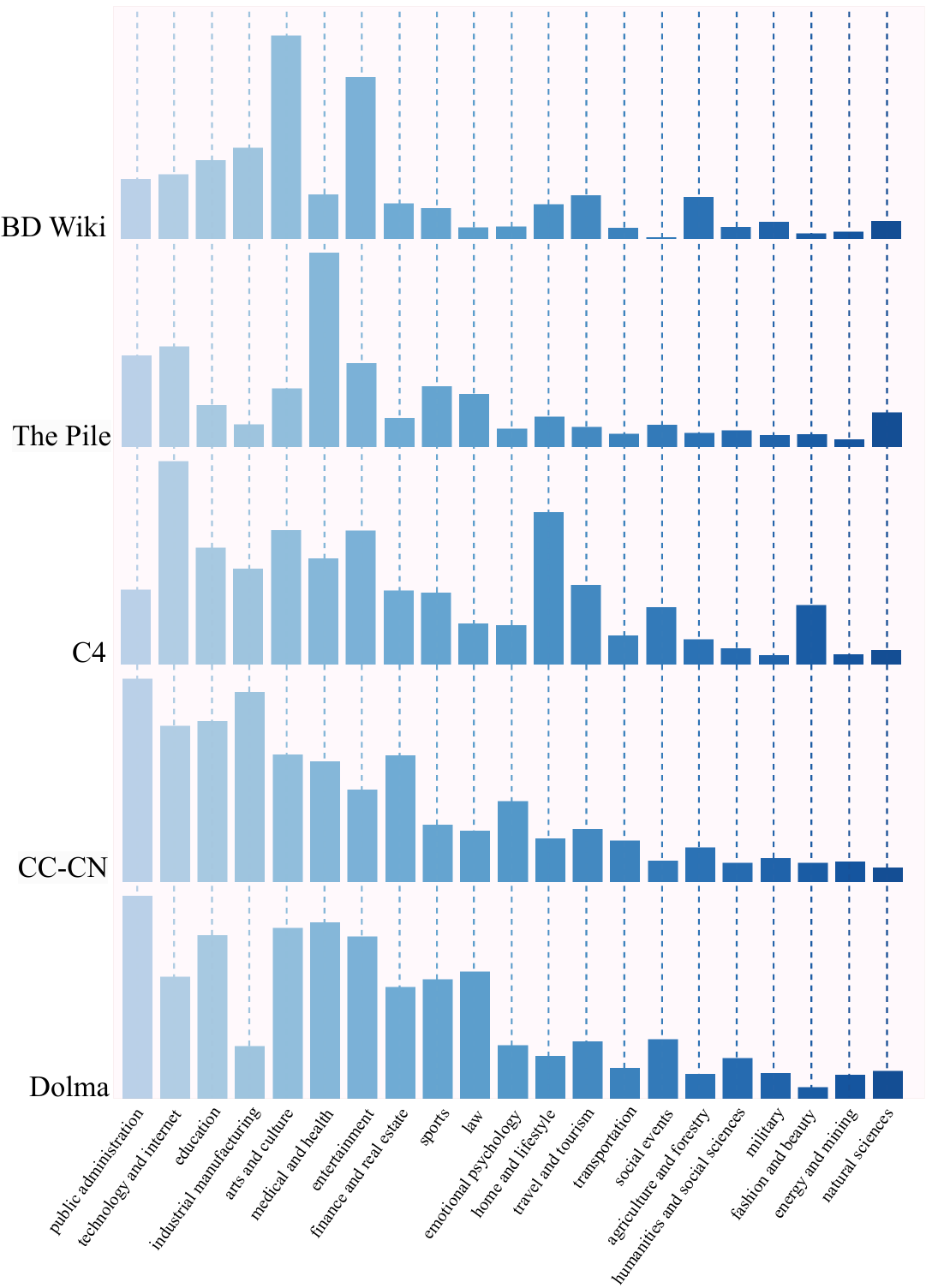}
    \caption{Distribution of first-level tags across different datasets, arranged in descending order by frequency in the decorated corpus.}
    \label{fig:enter-label}
\end{figure}

\begin{figure}[t]
    \centering
    \includegraphics[width=0.8\columnwidth]{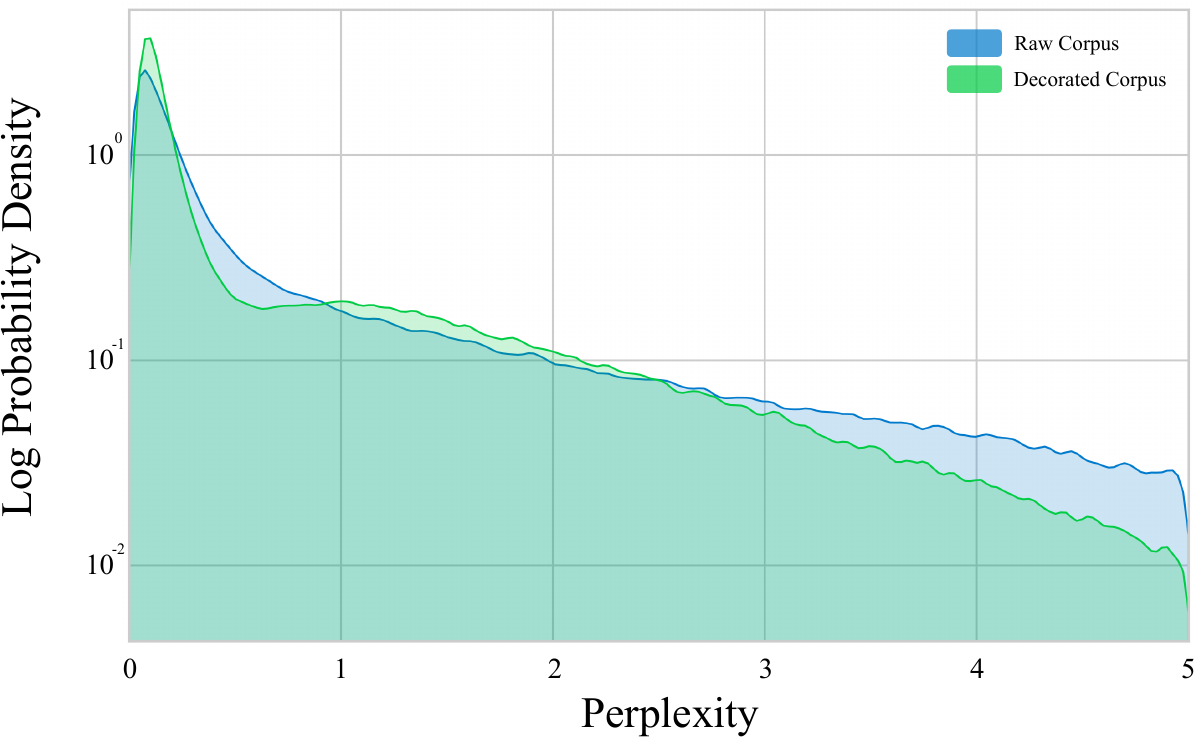}
    \caption{Perplexity distribution of the corpus.}
    \label{fig:ppl}
\end{figure}

\paragraph{Analysis.} We present the result of the tag tree in Figure~\ref{fig:taggingtree} and the word cloud of the tag tree in Figure~\ref{fig:wordcloud}.
To access the tag prediction performance, we manually re-annotated the existing validation split set with tags at the first, second, and third levels. We then compare the accuracy of DecorateLM and GPT-4 using this newly re-annotated validation set. As shown in Table~\ref{tab:taggingacc}, DecorateLM achieves performance comparable to that of GPT-4.
\subsection{Editing}
\begin{figure}[t]
    \centering
    \includegraphics[width=0.8\columnwidth]{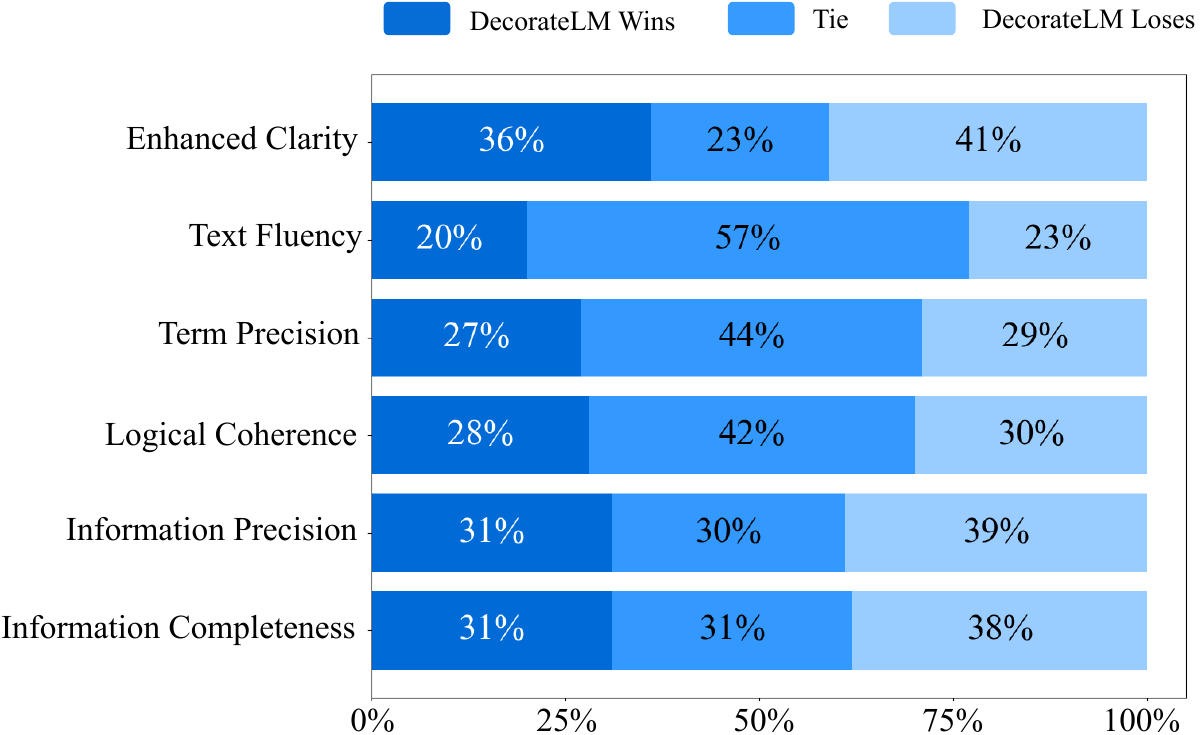}
    \caption{Human Preference for Edited Texts on Validation Set: DecorateLM vs. GPT-4.}
    \label{fig:editing human}
\end{figure}

The process of rating and tagging extracts valuable data from the pretraining corpus. Despite undergoing a rigorous cleaning pipeline, even high-quality data sourced from the internet may still retain some noise. Inspired by the work of~\citep{maini2024rephrasing}, we propose to enhance the utilization of this high-quality data by rephrasing it based on the intrinsic attributes of the samples. By transforming the data into different verbal forms, we aim to preserve the core information diversity of the pertaining stage while being as clean as the SFT-stage dataset. 
\paragraph{Annotated Dataset Construction.}
We begin by selecting 10,000 data samples, each containing between 50 and 2048 tokens, to create a noisy dataset. We observe that this noisy dataset continues to exhibit issues such as unclear expressions, lack of natural language fluency, and mixed topics that are not fully resolved by standard cleaning methods. 
This noisy dataset is rephrased using GPT-4 based on prompts in Appendix~\ref{editing prompt}.
\paragraph{Analysis.}
Due to the absence of a comprehensive metric for evaluating rephrased text against the original text, we design several custom metrics and use human evaluation to quality-check the rephrased texts. For each evaluation metric, we compare the rephrased outputs of DecorateLM and GPT-4, with human annotators rating each output as a win, lose, or tie. The evaluation metrics are as follows: \textit{Enhanced Clarity}, which determines the text's increased conciseness and clearer expression; \textit{Text Fluency}, which assesses the smoothness and readability of the text; \textit{Term Precision}, which checks the retention of specialized terminology; \textit{Logical Coherence}, which examines the consistency of causal and logical relationships within the text; \textit{Information Precision}, which verifies that the original meaning, core information, and arguments are accurately preserved; \textit{Information Completeness}, which ensures that no crucial information is missing from the text. 
The validation set size is 500. As presented in Figure~\ref{fig:editing human}, the editing model of DecorateLM, demonstrates satisfactory performance in this task.


\subsection{The Final Decorated Corpus}
After we train the DecorateLM on the curated annotated dataset, we proceed to decorate the pre-training corpus. 
Specifically, we select five large pre-training datasets including Common Crawl Chn (CC-CN), Dolma, C4, The Pile, and Baidu Wiki (BD-Wiki). 
Due to limited resources, we only sample a volume of 100 billion tokens from these datasets.

For the rated and tagged corpus, as shown in Figure~\ref{fig:dataset_level_quatlity}, the English datasets, Dolma and The Pile, exhibit relatively high ratings and low cross-entropy, making them relatively ideal training corpora that are high-quality and well-balanced across domains. In contrast, the Chinese datasets, BD-wiki and CC-CN, exhibit lower ratings and higher cross-entropy, indicating shortcomings in overall quality and data distribution. This also underscores the necessity of using DecorateLM to improve the quality of the non-English corpus. For the tagging result alone, the analysis of the distribution of these datasets across the first-level labels is illustrated in Figure ~\ref{fig:enter-label}.
Regarding the effectiveness of editing on the Decorated Corpus, the original and edited texts are assessed using the perplexity metric with the CCNet model~\citep{wenzek2019ccnet}. The results, shown in Figure ~\ref{fig:ppl}, indicate a significant reduction in perplexity following the editing process. This improvement suggests that the editing effectively organizes the data in a manner that is more conducive to learning by models, ensuring enhanced comprehensibility and learnability.


\begin{figure}[t]
    \centering
    \includegraphics[width=0.9\columnwidth]{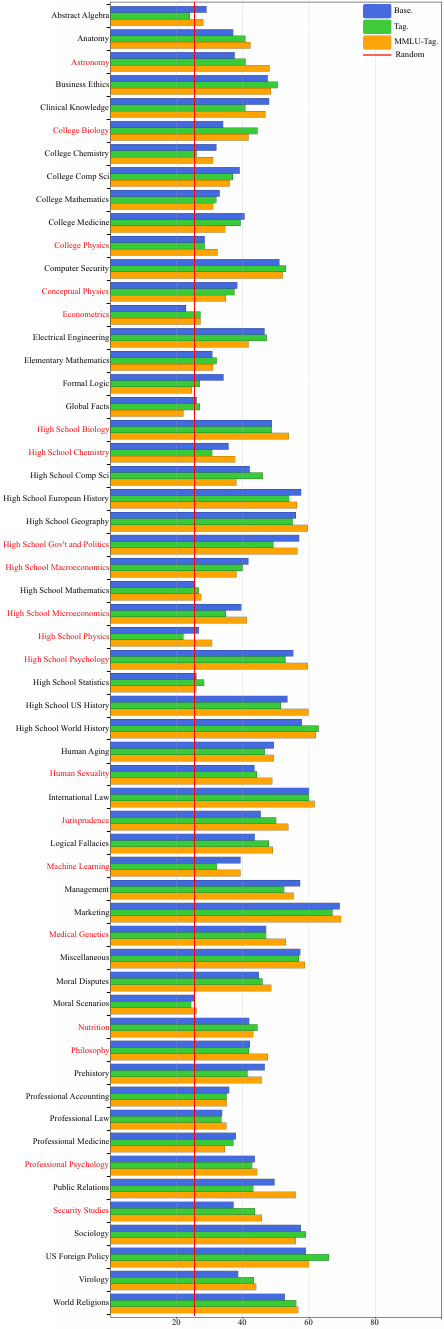}
    \caption{The performance of the MMLU-Tag. Model across the various subtasks of MMLU. The tasks where the sampling weights are increased on the corresponding tags based on the Tag. Methods are highlighted in red.}
    \label{fig:baseline vs taggingv1 vs tagging mmlu}
\end{figure}

\section{Experiments}
In this section, we conduct data experiments to demonstrate the effectiveness of decorated corpus.
\subsection{Experiment Setup}
We train the same SLM, MiniCPM-1.2B, used as the backbone for DecorateLM, aiming to improve its performance. MiniCPM-1.2B follows the multi-stage training pipeline~\citep{hu2024minicpm}. The stable training stage utilizes a constant learning rate until the decay stage, where the learning rate decreases rapidly. During the decay stage, the loss reduction accelerates significantly. This stage is deemed suitable for ablation studies on different data due to its substantial loss reduction and short training duration. We leverage the last checkpoint before the decay stage to reprocess the decay with both the raw and decorated corpora. Performance is evaluated against a wide range of publicly available benchmarks.

\subsection{Experiments on Rating}

Given the rating of each test sample, we can select each sample with a probability determined by these ratings~\citep {wettig2024qurating}.
We explore two sampling methods.

The first method, referred to as ``Separate Criterion Sampling'', follows the approach proposed by~\citep{wettig2024qurating}. Specifically, each criterion is given a weight that represents its relative importance. The sampling method begins from the criterion with the highest weight to the lowest one. The transition between criteria happens when the sampled data from the dimension satisfies its predetermined corpus proportion. Within each criterion, data is sampled according to the following weight~\ref{single dimension}. 
The ratings for the $i$-th data point in $t$-th criterion are calculated using the following equation: 
\begin{equation}
\label{single dimension}
W_{i,t} = e^{\frac{\text{score}_{i,t} - \lambda}{\tau}},
\end{equation}
where $i$ is the data point index and $t$ is the criterion index,  both $\lambda$ and $\tau$ are set to 50.


The second method, called ``Aggregate Criterion Sampling'', calculates the sampling weight $W_i$ for the $i$-th data as follows:
\begin{equation}
  W_i = \sum_{t=1}^{8} k_t \cdot e^{\frac{\text{score}_{t,i} - \mu_t}{\sigma_t}},
\end{equation}
where the parameter $k_t$ represents the relative significance of each rating dimension.

For both Rat.\ (Sep.) with weights and Rat.\ (Agg.) with $k_t$, the main method assigns a weight of 0.2 to the dimensions of Educational Value, Expertise, Fact and Trivia, and Reasoning Level, while the four remaining dimensions are each assigned a weight of 0.05  according to the authors' prior knowledge of the data quality. 

In practice, we sample 58.5B tokens but only use 45B tokens among them as the high-quality data. This has a similar effect as increasing the temperature of sampling in~\citep{wettig2024qurating}.
%


\begin{table*}[ht]
  \centering 
  \small
  \begin{tabular}{lccccccc}
    \toprule
    \textbf{Method} & \makecell{\textbf{C-Eval}\\ \textbf{(0-shot)}} & \makecell{\textbf{CMMLU}\\ \textbf{(5-shot)}} & \makecell{\textbf{AGI.}\\ \textbf{(5-shot)}} & \makecell{\textbf{MMLU}\\ \textbf{(5-shot)}} & \makecell{\textbf{Human.}\\ \textbf{(0-shot)}} & \makecell{\textbf{MBPP}\\ \textbf{(0-shot)}} & \makecell{\textbf{GSM.}\\ \textbf{(0-shot)}}\\
    \midrule
    Base.                       & 47.4           & 46.8         & 20.8                & 45.8            & 26.2          & 27.7              & 38.9          \\
    \midrule
    Tag.    & \scorechange{47.8}{\textuparrow0.4}{green}          & 46.8           & \scorechange{21.3}{\textuparrow0.5}{green}  &    \scorechange{47.3}{\textuparrow1.5}{green}          & \scorechange{27.4}{\textuparrow1.2}{green}             & \scorechange{28.4}{\textuparrow0.7}{green}          & \scorechange{40.0}{\textuparrow1.1}{green} \\ 
    \midrule
    Rat. (Sep.)    & \scorechange{45.2}{\textdownarrow2.2}{red}          & \scorechange{45.4}{\textdownarrow1.4}{red}           & \scorechange{26.4}{\textuparrow5.6}{green}  &    \scorechange{46.0}{\textuparrow0.2}{green}          & \scorechange{28.1}{\textuparrow1.9}{green}             & \scorechange{29.1}{\textuparrow1.4}{green}          & \scorechange{41.8}{\textuparrow2.9}{green}      
    \\
    Rat. (Agg.)    & \scorechange{\textbf{49.1}}{\textbf{\textuparrow1.7}}{green}          & \scorechange{47.0}{\textuparrow0.2}{green}           & \scorechange{26.3}{\textuparrow5.5}{green}  &    \scorechange{46.9}{\textuparrow1.1}{green}          & \scorechange{25.6}{\textdownarrow0.6}{red}             & \scorechange{30.3}{\textuparrow2.6}{green}          & \scorechange{42.5}{\textuparrow3.6}{green}
    \\
    Rat. (Agg.)\&Tag.                   & \scorechange{48.0}{\textuparrow0.6}{green}          & \scorechange{\textbf{47.9}}{\textbf{\textuparrow1.1}}{green}           & \scorechange{25.3}{\textuparrow4.5}{green}  &    \scorechange{46.0}{\textuparrow0.2}{green}          & \scorechange{28.7}{\textuparrow2.5}{green}             & \scorechange{28.1}{\textuparrow0.4}{green}          & \scorechange{40.9}{\textuparrow2.0}{green}          \\
    \midrule
    Edit.    & \scorechange{46.7}{\textdownarrow0.7}{red}          & \scorechange{47.1}{\textuparrow0.3}{green}           & \scorechange{23.8}{\textuparrow3.0}{green}  &    \scorechange{46.9}{\textuparrow1.1}{green}          & \scorechange{27.4}{\textuparrow1.2}{green}             & \scorechange{30.4}{\textuparrow2.7}{green}          & \scorechange{40.1}{\textuparrow1.2}{green} 
    \\
    Rat. (Agg.)\&Edit.    & \scorechange{48.1}{\textuparrow0.7}{green}          & \scorechange{47.8}{\textuparrow1.0}{green}           & \scorechange{\textbf{28.0}}{\textbf{\textuparrow7.2}}{green}  &    \scorechange{47.5}{\textuparrow1.7}{green}          & \scorechange{\textbf{31.7}}{\textbf{\textuparrow5.5}}{green}             & \scorechange{30.0}{\textuparrow2.3}{green}          & \scorechange{\textbf{42.6}}{\textbf{\textuparrow3.7}}{green}
    \\
    Rat. (Agg.)\&Tag.\&Edit.         & 47.4          & \scorechange{46.4}{\textdownarrow0.4}{red}           & \scorechange{24.3}{\textuparrow3.5}{green}  &    \scorechange{\textbf{47.6}}{\textbf{\textuparrow1.8}}{green}          & \scorechange{29.3}{\textuparrow3.1}{green}             & \scorechange{\textbf{30.9}}{\textbf{\textuparrow3.2}}{green}          & \scorechange{40.3}{\textuparrow1.4}{green}
    \\
    \toprule
    \textbf{Method} &  \makecell{\textbf{MATH}\\ \textbf{(4-shot)}} & \makecell{\textbf{BBH}\\ \textbf{(0-shot)}} & \makecell{\textbf{ARC-E}\\ \textbf{(0-shot)}} & \makecell{\textbf{ARC-C}\\ \textbf{(0-shot)}} & \makecell{\textbf{Trivia.}\\ \textbf{(0-shot)}} & \textbf{Avg. (DC)} & \textbf{Avg.} \\
    \toprule
    Base.                        & 3.5            & 28.5           & 78.2                & 61.8            & 6.0          & 37.5 & 36.1 \\
    \midrule
    Tag.    & \scorechange{4.6}{\textuparrow1.1}{green}          & \scorechange{27.8}{\textdownarrow0.7}{red}           & \scorechange{79.2}{\textuparrow1.0}{green}  &    \scorechange{62.1}{\textuparrow0.3}{green}          & \scorechange{12.7}{\textuparrow6.7}{green}             & \scorechange{41.8}{\textuparrow4.3}{green}          & \scorechange{37.5}{\textuparrow1.4}{green} 
    \\
    \midrule
    Rat. (Sep.)    & \scorechange{6.5}{\textuparrow3.0}{green}          & \scorechange{28.4}{\textdownarrow0.1}{red}           & \scorechange{78.8}{\textuparrow0.6}{green}  &    \scorechange{61.4}{\textdownarrow0.4}{red}          & \scorechange{10.4}{\textuparrow4.4}{green}             & \scorechange{39.2}{\textuparrow1.7}{green}          & \scorechange{37.4}{\textuparrow1.3}{green} 
    \\
    Rat. (Agg.)    & \scorechange{4.8}{\textuparrow1.3}{green}          & 28.5          & \scorechange{79.3}{\textuparrow1.1}{green}  &    \scorechange{\textbf{63.0}}{\textbf{\textuparrow1.2}}{green}          & \scorechange{15.6}{\textuparrow9.6}{green}             & \scorechange{41.1}{\textuparrow3.6}{green}          & \scorechange{38.5}{\textuparrow2.4}{green} 
    \\
    Rat. (Agg.)\&Tag.                  & \scorechange{\textbf{6.7}}{\textbf{\textuparrow3.2}}{green}          & \scorechange{28.0}{\textdownarrow0.5}{red}         & \scorechange{78.8}{\textuparrow0.6}{green}  &    \scorechange{62.6}{\textuparrow0.8}{green}          & \scorechange{13.7}{\textuparrow7.7}{green}             & \scorechange{43.1}{\textuparrow5.6}{green}          & \scorechange{38.3}{\textuparrow2.2}{green} 
    \\
    \midrule
    Edit.    & \scorechange{5.6}{\textuparrow2.1}{green}          & \scorechange{29.2}{\textuparrow0.7}{green}           & \scorechange{77.8}{\textdownarrow0.4}{red}  &    \scorechange{62.0}{\textuparrow0.2}{green}          & \scorechange{22.0}{\textuparrow16.0}{green}             & \scorechange{40.5}{\textuparrow3.0}{green}          & \scorechange{38.4}{\textuparrow2.3}{green} 
    \\
    Rat. (Agg.)\&Edit.    & \scorechange{4.3}{\textuparrow0.8}{green}          & \scorechange{\textbf{32.7}}{\textbf{\textuparrow4.2}}{green}           & \scorechange{\textbf{79.5}}{\textbf{\textuparrow1.3}}{green}  &    \scorechange{62.7}{\textuparrow0.9}{green}          & \scorechange{24.9}{\textuparrow18.9}{green}             & \scorechange{42.8}{\textuparrow5.3}{green}          & \scorechange{\textbf{40.2}}{\textbf{\textuparrow4.1}}{green} 
    \\
    Rat. (Agg.)\&Tag.\&Edit.          & \scorechange{5.5}{\textuparrow2.0}{green}          & \scorechange{29.8}{\textuparrow1.3}{green}           & \scorechange{77.9}{\textdownarrow0.3}{red}  &    \scorechange{\textbf{63.0}}{\textbf{\textuparrow1.2}}{green}          & \scorechange{\textbf{27.8}}{\textbf{\textuparrow21.8}}{green}             & \scorechange{\textbf{45.0}}{\textbf{\textuparrow7.5}}{green}          & \scorechange{39.6}{\textuparrow3.5}{green}
    \\
    \bottomrule
  \end{tabular}
  \caption{\label{tab:benchmark}
    Comparison of benchmark performance across different strategies.
  }
\end{table*}

\subsection{Experiments on Tagging}
We enhance the diversity and balance of different domains by incorporating a sampling strategy among tags. Intuitively, a large domain should be undersampled and a rare domain should be upsampled. 
Specifically, we sample an instance with a hierarchical tag of $a \rightarrow b\rightarrow c$ with the weight of 
\begin{equation}
\begin{split}
W_{a, b, c} = 
& \frac{N_{\mathrm{I}=a}^\alpha}{\sum_{i=1}^{N_{\mathrm{I}}} N_{\mathrm{I}=i}^\alpha} \cdot \frac{N_{\mathrm{I}=a, \mathrm{II}=b}^\beta}{\sum_{i=1}^{N_{\mathrm{I}=a, \mathrm{II}}} N_{\mathrm{I}=a, \mathrm{II}=i}^\beta} \cdot \\
& \frac{N_{\mathrm{I}=a, \mathrm{II}=b, \mathrm{III}=c}^\gamma}{\sum_{i=1}^{N_{\mathrm{I}=a, \mathrm{II}=b, \mathrm{III}}} N_{\mathrm{I}=a, \mathrm{II}=b, \mathrm{III}=i}^\gamma} ,
\end{split}
\label{eq: tag foumula}
\end{equation}
where $N_{{X}={x}}$ represents the number of instance whose belong to tag $x$ at tag level $X$. 
The exponents $\alpha, \beta, \gamma$ are similar to what is suggested by~\citep{lample2019cross} to tune the distribution to be smooth or concentrated.

For the combined method of Rat. (Agg) \& Tag. , we calculate the sampling weights by multiplying the weights of Rat. (Sep.) and Tag..

\textbf{Domain Coverage Criterion (Avg. (DC))}. To demonstrate the improvements brought by making the domain more balanced through tagging, we construct a domain coverage criterion by averaging the accuracy scores of 6 tasks within the following 5 domains. \textit{Sports} domain is represented by SportQA~\citep{xia2024sportqa} dataset. \textit{Medicine} domain is represented by MedMCQA~\citep{pal2022medmcqa} and MedQA-USMLE~\citep{jin2021disease} datasets. \textit{Law} domain is represented by JECQA~\citep{zhong2020jec} dataset. \textit{Natural sciences} domain is represented by SciQ~\citep{welbl2017crowdsourcing} dataset. \textit{Finance} domain is represented by OpenFinData dataset\footnote{\url{https://github.com/open-compass/OpenFinData}}.


\subsection{Experiments on Editing}

Building upon the existing methods (Baseline, Rat. (Agg.), and Rat. (Agg.)\&Tag.), we introduce the Editing approach. We randomly select one-third of the training data to be replaced with edited data.
\vspace{2pt} 
\subsection{Results}

In this section, we present the results of data experiments. Details and specific settings of the evaluation experiments can be found in Appendix~\ref{evaluation}.


As shown in Table~\ref{tab:benchmark}, the integration of various methods yields several significant insights:


\begin{itemize}
    \vspace{2pt} 
    \item \textbf{Rating:} Both rating sampling methods exhibit superior overall performance compared to the baseline. Rat.\ (Agg.) improves almost all tasks and achieves an overall average score increase of 2.4 points, which is greater than Rat.\ (Sep.).
    \vspace{2pt} 
    \item \textbf{Tagging:} The Tag. method shows a slight improvement over the baseline in overall benchmarks and achieves a significant 4.3-point increase on the Domain Coverage benchmark. The Rat.\ (Agg) \& Tag. method has comparable overall performance to Rat.\ (Agg), with an additional 2-point improvement on Avg.(DC). Moreover, to validate the effectiveness of domain filtering, we evaluate an MMLU-oriented tagging model, as depicted in Figure~\ref{fig:baseline vs taggingv1 vs tagging mmlu}. The model targets 20 specific MMLU subtasks, enhancing their sampling probability. It demonstrates improvement in 15 of these 20 tasks compared to the Tag. method, thereby affirming the efficacy of the tagging system in modifying domain composition for targeted reinforcement.
    \vspace{2pt} 
    \item \textbf{Editing:} Integration of the Editing method significantly enhances model performance on downstream tasks. Edit. increases the average score by 2.3 percentage points compared to the baseline, demonstrating its effectiveness in rephrasing training data.
    \vspace{2pt} 
    \item \textbf{Rating and Editing:} Rat.\ (Agg.)\&Edit. emerges as the best-performing method, enhancing the average score by 4.1 points relative to the baseline and demonstrating improvements across all tasks. Rat.\ (Agg.)\&Tag.\&Edit. attains the highest score on Avg. (DC) and maintains excellent performance in other tasks, suggesting that the integration of tagging with rating and editing expands the models' knowledge base without substantially compromising depth.

\end{itemize}

\vspace{5pt} 
\section{Conclusion}

In this paper, we present \textit{DecorateLM}, a \underline{d}ata \underline{e}ngineering method designed to refine the pretraining \underline{co}rpus through data \underline{ra}ting, \underline{t}agging and \underline{e}diting. \textit{DecorateLM} employs a dual-training strategy, wherein two student models with 1.2 B parameters are trained: one designed for rating and tagging, and the other focused on editing. Our experiments show that introducing rating and editing in data corpus significantly enhances data quality by improving the overall performance of SLM on various existing benchmarks. Furthermore, our empirical study verifies that the implemented tagging strategy achieves a more balanced distribution of categories within the training dataset. This equilibrium in categorization enables a more thorough comprehension of SLM proficiency across diverse domains. These encouraging results underscore the importance of training data quality in fully exploiting the capabilities of Large Language Models, thereby suggesting several compelling avenues for future research.

\newpage
\section{Limitations}

Our study, while enhancing the quality of data effectively, is subject to several limitations.
Firstly, the biases present in GPT-4 may be reflected in the fine-tuning data used for DecorateLM, potentially causing DecorateLM to inherit these biases
Additionally, due to computational and time constraints, we limit our model training to 1.2 billion parameter models using high-quality data. The generalizability of our findings would benefit from replication with larger language models and a wider range of datasets.
Thirdly, our investigation is confined to training models during the decay stage using the Decorated Corpus. An additional dimension to our work would involve creating a dataset of 1.1 trillion tokens with DecorateLM, followed by training a model from scratch on this enlarged dataset, which we believe represents an important direction for future research. 

Moreover, although DecorateLM performs well in filtering data from large-scale web data, its ability to handle more specialized domains still requires improvement. The classification and labeling of the diverse content of the real world by humans are challenging to fully capture with a three-layer labeling system. Future research could explore a more granular labeling system to enhance the model's precision and breadth in professional fields.
Lastly, while DecorateLM considered both English and Chinese, it did not take other languages such as French and Russian into account, which may limit its generalizability to other languages.

An additional limitation lies in the current approach to sampling, which may not adequately capture the nuanced relationships between ratings and taggings across various tasks. Therefore, future research should explore a wider array of sampling strategies for rating and tagging to assess their impact on task performance more comprehensively. 

\section{Ethical Considerations}

As we develop DecorateLM, we recognize the inherent risk of introducing or magnifying biases within our datasets. The training process, while intended to refine and improve data accuracy, could inadvertently perpetuate biases present in the original data. This raises significant ethical concerns, as biased data can lead to unfair outcomes in decision-making processes that rely on our enhanced training data.


\bibliography{custom}

\appendix

\newpage
\section{Full Prompts}
\subsection{Prompts of Rating}
\label{rating prompt}

\begin{tcolorbox}[title = {Prompt Template},breakable]
\small
Compare which text \{criterion\} \\
Your judgement should not be influenced by the language the text is written in, the length of the text and the order in which the texts are presented. \\
If the texts have similar quality, you should still make a relative judgement and choose the label of the preferred text.  \\
You must respond with format: \\
"Choice: 1 or 2\textbackslash{n}Why: reason of choice"  \\ \\
Text 1:
... \{text\_1\} ...\\ \\
Text 2:
... \{text\_2\} ... \\ \\
Now you have to choose between either 1 or 2. Note that respond only with the format mentioned.
\end{tcolorbox}

\begin{tcolorbox}[title = {Educational Value},breakable]
\small
has more educational value. It has more educational value if it includes clear explanations, step-by-step reasoning, or detailed concepts which is clear enough for children to understand. \\
Prefer text which has more detailed ideas or explanations which is sufficiently clear to convey them to a child.
\end{tcolorbox}

\begin{tcolorbox}[title = {Expertise},breakable]
\small
requires greater expertise and deeper prerequisite knowledge to understand it. \\
For example, "The relativistic Dirac equation, which combines principles of quantum mechanics and special relativity, predicts the existence of antimatter and elucidates the intrinsic spin of fundamental particles." requires great physics expertise to understand.
\end{tcolorbox}

\begin{tcolorbox}[title = {Fact and Trivia},breakable]
\small
contains more facts and trivia. The facts and trivia should be accurate. \\
Prefer text which have more number of facts. Put lower priority to facts which contain mathematical calculations and with too deep "concepts and explanations" .
\end{tcolorbox}

\begin{tcolorbox}[title = {Reasoning Level},breakable]
\small
has higher reasoning level. It has high reasoning level when it requires more reasoning, logical and mathematical thinking skills or chain of thought thinking.
\end{tcolorbox}

\begin{tcolorbox}[title = {Scarcity},breakable]
\small
is more relatively unknown. It should be truthful and little known to the general public. \\
Prefer unpopular accurate facts over fictional stories.
\end{tcolorbox}

\begin{tcolorbox}[title = {Structural Format},breakable]
\small
has better structural format. It has better structural format when it has a well-defined structure such as outline format, Markdown, numbered list, bulleted list, JSON, table format, headings and subheadings format or other organizational templates. \\
First, consider the visual structure of text. Then, only consider the content or logical flow of text.
\end{tcolorbox}

\begin{tcolorbox}[title = {Story-likeness},breakable]
\small
is more likely to be a story. It is more like a story when it narrates a story or it describes a scene or situation in details.
\end{tcolorbox}

\begin{tcolorbox}[title = {Subjectivity},breakable]
\small
contains more subjectivity, e.g, it includes more subjective perspectives, opinions, personal views or feelings. Avoid choosing text which conveys objective, factual and widely accepted, accurate knowledge. \\
Prefer text which personal opinions such as dialogues or feelings over text which seems like a formal examination question and answer.
\end{tcolorbox}

\begin{tcolorbox}[title = {Generate Structural Format Data},breakable]
\small
You are tasked with generating text data that has clear and organized formatting structures. Some structural format are list, markdown, headings and subheadings, table, json, html, xml, latex, columnar formats etc. \\
The data should maintain a coherent structure with organized sections, numbering, tables, code formatting, hierarchical structure, outlines or other organizational templates where appropriate. You should not include all of the formats in one data. One data can mix of one, two or three formats. \\
You can add various knowledge and facts into data to make data more informative and longer. \\
Please generate 3 lengthy and informative examples about `<topic>` showcasing different formatting styles and content. Split examples with <split>
\end{tcolorbox}

\subsection{Prompts of Tagging}
\label{tagging prompt}
\begin{tcolorbox}[title = {Prompt Template For Summary},breakable]
\label{summary}
\small
Your objective is to summarize the provided text: [begin] \{instance\} [end], within 100 words, including the relevant information for the use case in the summary as much as possible. \\
    The summary will represent the input data for clustering in the next step. \\
    Be concise and clear. \\
    Do not add phrases like "This is the summary of" or "Summarized text:"... \\
    Do not include any line breaks in the summary. \\
    Provide your answer in English only. \\
    Your comprehensive output should mirror this structure: \{\{"summary": ""\}\}. \\
\end{tcolorbox}

\begin{tcolorbox}[title = {Prompt Template For First-level Tagging},breakable]
\label{first-level tagging}
\small
You are an advanced tagging system designed to identify the most pertinent theme within a given text passage: [begin] \{instance\} [end].\\
    Your role is to analyze the text meticulously and choose the most fitting tag from the predefined list: Natural Sciences, Humanities and Social Sciences, Industrial Manufacturing, Medical and Health, Agriculture and Forestry, Energy and Mining, Finance and Real Estate, Education, Transportation, Technology and Internet, Law, Military, Travel and Tourism, Entertainment, Arts and Culture, Emotional Psychology, Fashion and Beauty, Sports, Home and Lifestyle, Public Administration, and Social Events.\\
    Your task is to determine the single most relevant tag that encapsulates the primary theme of the text. \\
    Your selection should be substantiated with a detailed explanation, elucidating why this tag is the most accurate representation of the text's central subject matter.\\
    Your output should follow this structure: \{\{"tag": "Selected Tag", "explanation": "Provide a detailed explanation in English  on why this is the most fitting tag."\}\}.
\end{tcolorbox}

\begin{tcolorbox}[title = {Prompt Template For Second-level And Third-level Tagging},breakable]
\label{second-level and third level tagging}
\small
You are an advanced tagging system designed to categorize a given text passage related to the first level tag "\{first\_level\_tag\}" into specific second and third-level tags within a predefined hierarchy. \\
    Here is the tag hierarchy for the "\{first\_level\_tag\}" category in json format: \{tag\_tree\} \\
    Here is the given text passage: [begin] \{instance\} [end]. \\
    Your task is to analyze the text snippet above and assign the most fitting second-level and third-level tags, ensuring both tags align within the same hierarchical path. \\
    The output should precisely reflect the main focus of the text, justifying why these tags are the most suitable choices. \\
    Your output should follow this structure: \{\{"second\_level\_tag": "Selected Second Level Tag", "third\_level\_tag": "Selected Third Level Tag", "explanation": "Provide a detailed explanation in English on why these tags accurately represent the text's core content."\}\}.
\end{tcolorbox}

\subsection{Prompts of Editing}
\label{editing prompt}

\begin{tcolorbox}[title = {Editing Template},breakable]
\small
For the following paragraph give me a diverse paraphrase of the same in high quality language as in sentences on Wikipedia. Generate text directly from the provided content. Do not exceed the original information or add explanations. \\
text:
\end{tcolorbox}

\section{DecorateLM Training}
\subsection{Details of rating and tagging model}
\label{decoratelm training details rt}
We employ MiniCPM-1.2B~\citep{hu2024minicpm} as our base model. Utilizing the previously proposed rating and tagging methodologies, we collect rating and three-level tagging of 30,000 training data samples and subsequently apply supervised fine-tuning to the MiniCPM-1.2B with a learning rate of 0.00125 and total batch size of 480 every iteration. The fine-tuning process is conducted on three machines, each equipped with eight Nvidia A100 GPUs. We implement an decay step every 120 iterations and a warm-up phase of 3 iterations, yielding distilled rating and tagging models. We observe that only 200 steps are needed to fine-tune the model to its optimal performance in rating and tagging.

\subsection{Details of editing model}
\label{decoratelm training details re}
Similar to the rating and tagging model, we utilize the previously proposed editing method and collect 10,000 data samples with rephrased content by GPT-4. Subsequently, we apply supervised fine-tuning to MiniCPM-1.2B with the same method and hyperparameters as the rating and tagging model, yielding an editing model. We observe that fine-tuning the model for optimal performance in editing tasks requires 600 steps, a notably higher number compared to the steps needed for the rating and tagging model. This increased demand for training iterations likely reflects the greater complexity and difficulty associated with editing tasks.

\section{Further Analysis of DecorateLM}
\subsection{Cost Analysis}
Utilizing the vLLM framework~\citep{kwon2023efficient} and Ray~\citep{moritz2018ray}, we facilitate the generation of synthetic data across distinct phases with varying processing efficiencies on a single Nvidia A100 GPU. In the rating and tagging phase, the MiniCPM-1.2B model processes 16 million tokens per hour, requiring approximately 6,250 GPU hours to generate 100 billion tokens. Conversely, in the editing phase, the same model configuration processes 12.5 million tokens per hour, necessitating around 8,000 GPU hours for the production of an equivalent volume of tokens. 

\subsection{Details of Decorated Corpus}
\begin{figure*}[t]
    \centering
    \includegraphics[width=0.8\textwidth]{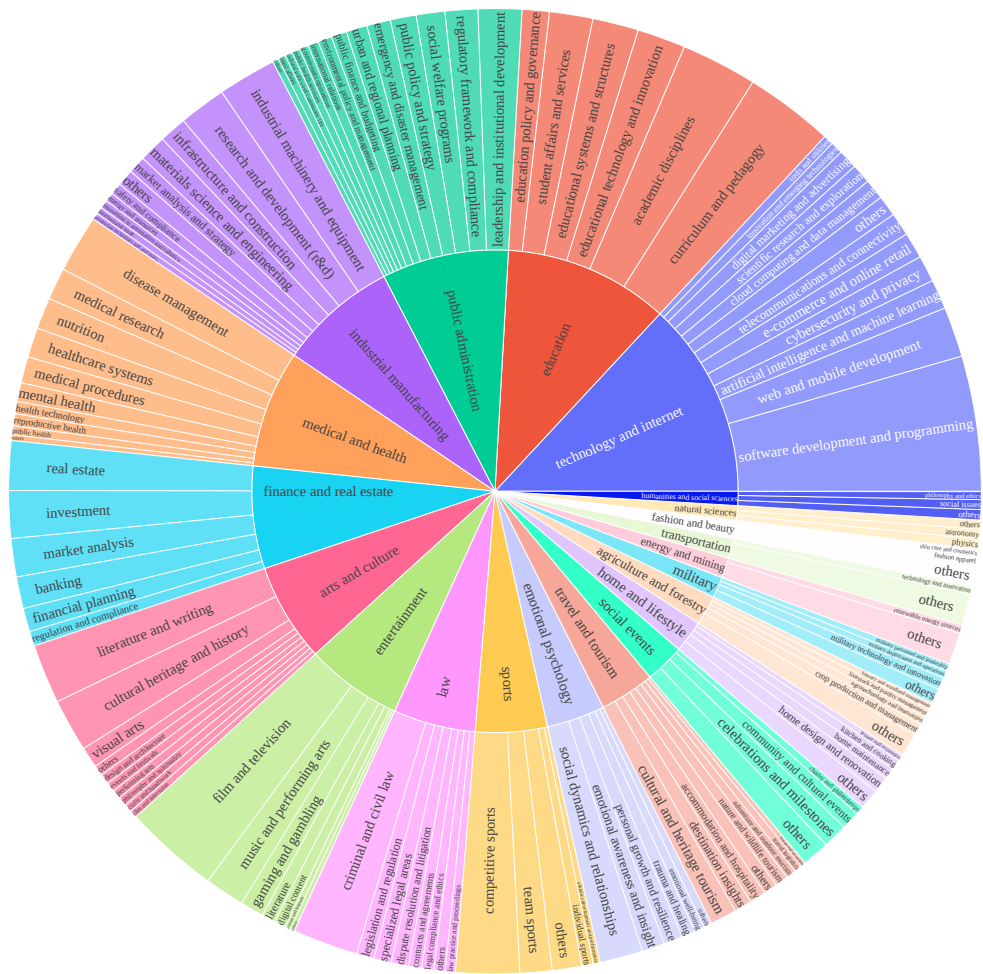}
    \caption{The tagging tree hierarchy. Only first and second-level tags are shown.}
    \label{fig:taggingtree}
\end{figure*}

The Decorated Corpus is constructed from a variety of datasets, each contributing to the total composition according to the proportions specified in Table~\ref{tab:decorated_corpus}.
\paragraph{Dolma.}
Dolma dataset~\citep{soldaini2024dolma} encompasses a comprehensive corpus designed for advancing the field of language model pretraining.
\paragraph{CC-CN.} CC-CN dataset is composed of a combination of sources from~\citep{CLUECorpus2020},~\citep{wei2023skywork}, and~\citep{wu2021yuan}
\paragraph{C4.} C4 dataset~\citep{raffel2020exploring} represents a significant milestone in the field of natural language processing, particularly within the domain of transfer learning.
\paragraph{The Pile.} The Pile dataset~\citep{gao2020pile} is a substantial contribution to large-scale language model training, featuring an extensive corpus of 825 GiB of English text.
\paragraph{BD Wiki.} The BD Wiki dataset, derived from the Baidu Baike\footnote{\url{https://baike.baidu.com/}}, is a semi-open Chinese online encyclopedia operated by Baidu Inc.

\section{Training With Decorated Corpus}
\label{evaluation}
\subsection{Experimental Details}
We employ the pre-decay version of MiniCPM-1.2B, pre-trained on a corpus comprising 800 billion tokens, as our base model. For training, the Decorated Corpus and additional high-quality datasets are utilized. The base model undergoes a decay process over 20,000 steps with a learning rate of 0.01 and a batch size of 1200 tokens per iteration, distributed across 10 machines, each equipped with eight A100-80GB GPUs. A decay step is implemented every 5000 iterations.
\begin{table*}[ht]
  \centering
  \begin{tabular}{lccccc}
    \toprule
    \textbf{Dataset} & \textbf{Dolma} & \textbf{CC-CN} & \textbf{C4} & \textbf{The Pile} & \textbf{BD Wiki} \\
    \midrule
    \# Tokens (millions) & 320 & 290 & 200 & 100 & 90 \\
    \bottomrule
  \end{tabular}
  \caption{\label{tab:decorated_corpus}
    Composition of the Decorated Corpus Dataset.
  }
\end{table*}

\begin{table*}[ht]
  \centering
  \small
  \begin{tabular}{lccccccc}
    \toprule
    \textbf{Method} & \makecell{\textbf{Sport.}\\ \textbf{(0-shot)}} & \makecell{\textbf{MedMC.}\\ \textbf{(0-shot)}} & \makecell{\textbf{Med.-US.}\\ \textbf{(0-shot)}} & \makecell{\textbf{JEC.}\\ \textbf{(0-shot)}} & \makecell{\textbf{SciQ}\\ \textbf{(0-shot)}} & \makecell{\textbf{OpenFin.}\\ \textbf{(0-shot)}} & {\textbf{Avg. (DC)}} \\
    \midrule
    Base.                       & 16.5           & 29.8    & 28.0                & 31.4            & 71.3          & 48.1              & 37.5          \\
    \midrule
    Tag.    & \scorechange{20.9}{\textuparrow4.4}{green}          & \scorechange{36.9}{\textuparrow7.1}{green}           & \scorechange{34.4}{\textuparrow6.4}{green}  &    \scorechange{35.4}{\textuparrow4.0}{green}          & \scorechange{74.0}{\textuparrow2.7}{green}             & \scorechange{\textbf{48.9}}{\textbf{\textuparrow0.8}}{green}          & \scorechange{41.8}{\textuparrow4.3}{green} \\ 
    \midrule
    Rat. (Sep.)    & \scorechange{7.0}{\textdownarrow9.5}{red}          & \scorechange{36.8}{\textuparrow7.0}{green}           & \scorechange{36.6}{\textuparrow8.6}{green}  &    \scorechange{35.4}{\textuparrow4.0}{green}          & \scorechange{77.2}{\textuparrow5.9}{green}             & \scorechange{42.3}{\textdownarrow5.8}{red}          & \scorechange{39.2}{\textuparrow1.7}{green}      
    \\
    Rat. (Agg.)    & \scorechange{15.0}{\textdownarrow1.5}{red}          & \scorechange{36.9}{\textuparrow7.1}{green}           & \scorechange{37.1}{\textuparrow9.1}{green}  &    \scorechange{34.5}{\textuparrow3.1}{green}          & \scorechange{77.4}{\textuparrow6.1}{green}             & \scorechange{45.7}{\textdownarrow2.4}{red}          & \scorechange{41.1}{\textuparrow3.6}{green}
    \\
    Rat. (Agg.)\&Tag.                    & \scorechange{22.2}{\textuparrow5.7}{green}          & \scorechange{\textbf{39.9}}{\textbf{\textuparrow10.1}}{green}           & \scorechange{36.3}{\textuparrow8.3}{green}  &    \scorechange{36.4}{\textuparrow5.0}{green}          & \scorechange{78.4}{\textuparrow7.1}{green}             & \scorechange{45.2}{\textdownarrow2.9}{red}          & \scorechange{43.1}{\textuparrow5.6}{green}
    \\
    \midrule
    Edit.    & \scorechange{16.8}{\textuparrow0.3}{green}          & \scorechange{33.0}{\textuparrow3.2}{green}           & \scorechange{32.1}{\textuparrow4.1}{green}  &    \scorechange{\textbf{36.6}}{\textbf{\textuparrow5.2}}{green}          & \scorechange{75.9}{\textuparrow4.26}{green}             & \scorechange{48.7}{\textuparrow0.6}{green}          & \scorechange{40.5}{\textuparrow3.0}{green} 
    \\
    Rat. (Agg.)\&Edit.    & \scorechange{17.5}{\textuparrow1.0}{green}          & \scorechange{36.9}{\textuparrow7.1}{green}           & \scorechange{39.5}{\textuparrow11.5}{green}  &    \scorechange{36.5}{\textuparrow5.1}{green}          & \scorechange{80.5}{\textuparrow9.2}{green}             & \scorechange{45.6}{\textdownarrow2.5}{red}          & \scorechange{42.8}{\textuparrow5.3}{green}
    \\
    Rat. (Agg.)\&Tag.\&Edit.          & \scorechange{\textbf{25.8}}{\textbf{\textuparrow9.3}}{green}          & \scorechange{38.8}{\textuparrow9.0}{green}           & \scorechange{\textbf{40.1}}{\textbf{\textuparrow12.1}}{green}  &    \scorechange{36.4}{\textuparrow5.0}{green}          & \scorechange{\textbf{80.7}}{\textbf{\textuparrow9.4}}{green}             & 48.1          & \scorechange{\textbf{45.0}}{\textbf{\textuparrow7.5}}{green}          \\
    \bottomrule
  \end{tabular}
  \caption{\label{tab:domain benchmark}
    Comparison of rare domain benchmark performance across different strategies.
  }
\end{table*}
\subsection{Evaluation Details}

The overall evaluation utilizes the open-source tool UltraEval\footnote{\url{https://ultraeval.openbmb.cn/home}}. The underlying
inference and acceleration use the open-source framework vLLM~\citep{kwon2023efficient}, and the dataset includes commonly used datasets: C-Eval~\citep{huang2024c} and CMMLU~\citep{li2023cmmlu} for Chinese knowledge, AGI-Eval~\citep{zhong2023agieval} for World Knowledge, MMLU~\citep{hendrycks2020measuring} for English knowledge, HumanEval~\citep{chen2021evaluating} and MBPP~\citep{austin2021program} for coding, GSM8K~\citep{cobbe2021training} and MATH~\citep{hendrycks2021measuring} for mathematics, and BBH~\citep{srivastava2022beyond} for logic reasoning, and ARC-E~\citep{clark2018think}, ARC-C~\citep{clark2018think}for commonsense reasoning, and TriviaQA~\citep{joshi2017triviaqa} for Reading Comprehension. Additionally, we conduct the Domain Coverage (DC) benchmark to evaluate the model's capability across various domain-specific knowledge bases. The DC Benchmark includes datasets such as SportQA~\citep{xia2024sportqa} for sports, MedMCQA~\citep{pal2022medmcqa} and MedQA-USMLE~\citep{jin2021disease} for medicine, JECQA~\citep{zhong2020jec} for law, SciQ~\citep{welbl2017crowdsourcing} for natural sciences, and OpenFinData\footnote{\url{https://github.com/open-compass/OpenFinData}} for finance.

\section{Inspecting cases of DecorateLM}
\label{case_study}
\begin{tcolorbox}[title = {Case 1}, colframe=darkgreen, colback=lightgray, breakable]
\small
\textbf{Text:}\\Photodynamic diagnosis (PDD) is available at some centres. PDD makes cancer cells visible under violet light. This makes abnormal growths (tumours) easier to find and remove. More complete removal reduces the risk of cancer coming back.PDD is used in urology to test for bladder cancer. A catheter is inserted through the urethra and into the bladder. A solution of 5-aminolaevulinic acid or hexaminolaevulinic acid is passed through the catheter into the bladder. Then the catheter is taken out immediately. Cancer cells in the bladder process the chemicals in the solution and become fluorescent under violet light. No side-effects or complications have been reported for PDD.\\
\textbf{Rating:}
\\
\begin{tabular}{@{}llll}
Educational Value: & 99 & Expertise: & 92 \\ 
Fact and Trivia: & 92 & Scarcity: & 85 \\
Reasoning Level: & 93   & Story-likenss: & 55 \\ 
Structural format: & 91 & Subjectivity: & 25 \\
\end{tabular}
\\
\textbf{Tagging:}
\\
\begin{tabular}{@{}ll}
First Level Tag: & Medical And Health \\
Second Level Tag: & Medical Procedures \\
Third Level Tag: & Diagnostic Procedures \\
\end{tabular}
\\
\textbf{Editing:}
\\
Photodynamic diagnosis (PDD) is accessible at certain facilities, a technique that enables the visualization of cancer cells under violet light, thereby enhancing the detection and removal of irregularities, such as tumours, with increased precision. This method further minimizes the likelihood of cancer cells recurring. In the field of urology, PDD is utilized for the identification of bladder cancer. The procedure involves the insertion of a catheter through the urethra into the bladder. Within the catheter, a solution containing 5-aminolaevulinic acid or hexaminolaevulinic acid is delivered. Following this, the catheter is promptly removed. It is observed that cancer cells within the bladder respond to these chemicals, converting them into fluorescent particles under the influence of violet light. Due to its safe and effective operation, no adverse effects or issues have been noted associated with PDD.
\end{tcolorbox}

\begin{tcolorbox}[title = {Case 2}, colframe=darkgreen, colback=lightgray, breakable]
\small
\textbf{Text:}\\Tron (TRX) is an ambitious project which aims to own the bragging rights regarding the entertainment industry. To make that a reality, a day hardly ends...Tron (TRX) is undoubtedly an ambitious project in the crypto community, and it aims to own the bragging rights of the entertainment industry. However, they are...Before I start explaining how Tron (TRX) Mainnet, Odyssey, is enhancing the chances of Tron adoption already, I need to explain some crucial thoughts with you....What will be the price of Tron (TRX) in five years from now? Tron is one of the top cryptocurrencies in the world (always in the spotlight). The cryptocurrency’s mainnet is now live, and on its independence day, it...Pornhub, one of the largest X-rated video sites, expands its virtual currency payment options announcing they will be soon accepting Tron (TRX). In recent times, Tron...Can Tron (TRX) become one of the top 5 cryptos in the market now?In recent days, Tron (TRX) has been experiencing one of the biggest tractions of the market, a fact that without a doubt is helping the token...\\
\textbf{Rating:}
\\
\begin{tabular}{@{}llll}
Educational Value: & 45 & Expertise: & 54 \\ 
Fact and Trivia: & 80 & Scarcity: & 10 \\
Reasoning Level: & 51   & Story-likenss: & 60 \\ 
Structural format: & 55 & Subjectivity: & 80 \\
\end{tabular}
\\
\textbf{Tagging:}
\\
\begin{tabular}{@{}ll}
First Level Tag: & Technology and Internet \\
Second Level Tag: & Entertainment and Media \\
Third Level Tag: & Digital Media \\
\end{tabular}
\\
\textbf{Editing:}
\\
Tron (TRX) embodies a visionary endeavor to dominate the boundaries of the entertainment sector, a goal that promises to be achieved on a nearly perpetual basis. In pursuit of this ambitious objective, Tron (TRX) continually strives to enhance its presence. In the realm of cryptocurrency, Tron (TRX) stands out for its bold ambitions, vying for control over the entertainment industry's prestige. Despite its aspirations, Tron (TRX) is already making strides in securing a larger user base, thanks to its innovative Mainnet, Odyssey. This initiative is pivotal in laying the groundwork for Tron's broader adoption. As we reflect on the potential future value of Tron (TRX) in the coming five years, it is noteworthy that Tron holds a prominent position among the world's cryptocurrencies, drawing significant attention. The cryptocurrency's flagship network has commenced operations, and its founding day marks an important milestone for its development. The entertainment sector's leading platform, Pornhub, announces a expansion into virtual currency payments, revealing plans to begin accepting Tron (TRX) as a payment method shortly. This move highlights the growing acceptance and adoption of Tron (TRX) within the digital landscape. Furthermore, there exists a discussion surrounding the potential for Tron (TRX) to surpass other top 5 cryptos in the market, with recent market trends clearly indicating its robust growth and widespread acceptance.
\end{tcolorbox}

\begin{tcolorbox}[title = {Case 3}, colframe=darkgreen, colback=lightgray, breakable]
\small
\textbf{Text:}
\\
Gown, \$12,900, Valentino, Bal Harbour Shops and Design District; diamond stud earrings, price upon request, elanjewels.us.
Jennifer Hudson, emotion is everything. It’s how she breathes life into a character. It’s how she makes a song explode. And it’s why—since Hudson was a child—people are drawn to her talent like a moth to a flame. Well, wait until you see her newest film.
Larkin coat, \$6,770, by Erdem at Saks Fifth Avenue, Bal Harbour Shops, Brickell City Centre and Dadeland Mall; satin Bullet bodysuit, \$350, by Fleur du Mal at Intermix, Bal Harbour Shops, Brickell City Centre and Lincoln Road; Kimmy belt, \$625, at Isabel Marant, Design District; printed velvet trousers, \$900, by Paco Rabanne at The Webster, Bal Harbour Shops and South Beach; Ellabrita strass sandal 105, \$1,150, by René Caovilla at Neiman Marcus, Bal Harbour Shops and Shops at Merrick Park; diamond earrings, price upon request, at elanjewels.us.
Gown, \$25,000, Valentino, Bal Harbour Shops and Design District; feather boa, \$3,990, Loewe, Design District.
\\
\textbf{Rating:}
\\
\begin{tabular}{@{}llll}
Educational Value: & 10 & Expertise: & 2 \\ 
Fact and Trivia: & 51 & Scarcity: & 1 \\
Reasoning Level: & 11   & Story-likenss: & 50 \\ 
Structural format: & 36 & Subjectivity: & 63 \\
\end{tabular}
\\
\textbf{Tagging:}
\\
\begin{tabular}{@{}ll}
First Level Tag: & Fashion and Beauty \\
Second Level Tag: & Fashion Apparel \\
Third Level Tag: & Formal Attire \\
\end{tabular}
\\
\textbf{Editing:}
\\
A gown from Valentino, Bal Harbour Shops and Design District, retails for \$12,900. Additionally, diamond stud earrings are available at a custom price, with the source being elanjewels.us. Jennifer Hudson is celebrated for her profound emotional depth, capturing the essence of her characters and elevating musical pieces to new heights. Her innate talent, which has attracted a multitude of fans since she was a young age, is poignantly depicted in her latest cinematic venture.
Erdem's Larkin coat, priced at \$6,770, is showcased at Saks Fifth Avenue, alongside other collections in Bal Harbour Shops, Brickell City Centre, and Dadeland Mall. The Bullet bodysuit, priced at \$350, features a satin material by Fleur du Mal at Intermix, along with other merchandise in these same locations. A belt by Kimmy, priced at \$625, is available at Isabel Marant in the Design District. For a more contemporary look, a printed velvet trouser, priced at \$900, by Paco Rabanne is offered at The Webster in Bal Harbour Shops and South Beach.
Elenabrita's Ellabrita strass sandal 105, priced at \$1,150, is designed by René Caovilla and available at Neiman Marcus, Shops at Merrick Park, and additional retailers. Diamond earrings, once requested, can be purchased from elanjewels.us.
A gown from Valentino, priced at \$25,000, is available from Bal Harbour Shops and Design District, while a feather boa, priced at \$3,990, adds a distinctive touch to Loewe's designs in the Design District.
\end{tcolorbox}

\end{document}